\documentclass[10pt,twocolumn,letterpaper]{article}

\usepackage{cvpr}
\usepackage{times}
\usepackage{epsfig}
\usepackage{graphicx}
\usepackage{amsmath}
\usepackage{amssymb}
\usepackage{booktabs}
\usepackage{subfig}

\usepackage[round]{natbib}
\usepackage[breaklinks=true,bookmarks=false]{hyperref}

\cvprfinalcopy 

\def\cvprPaperID{****} 

\newcommand*{\figuretitle}[1]{
    {\centering
    \textbf{#1}
    \par\medskip}
}

\begin{document}

\title{Proxy Datasets for Training Convolutional Neural Networks}

\author{Sam Shleifer\\
{\tt\small sshleifer [AT] stanford.edu }
\and
Eric Prokop\\
{\tt\small ecprokop [AT] stanford.edu}
}

\maketitle

\begin{abstract}
   One of the biggest bottlenecks in a machine learning workflow is waiting for models to train. Depending on the available computing resources, it can take days to weeks to train a neural network on a large dataset with many classes such as ImageNet. For researchers experimenting with new algorithmic approaches, this is impractically time-consuming and costly. We aim to generate smaller ``proxy datasets" where experiments are cheaper to run but results are highly correlated with experimental results on the full dataset. 
   We generate these proxy datasets using by randomly sampling from examples or classes, training on only the easiest or hardest examples and training on synthetic examples generated by ``data distillation". We compare these techniques to the more widely used baseline of training on the full dataset for fewer epochs. For each proxying strategy, we estimate three measures of ``proxy quality": how much of the variance in experimental results on the full dataset can be explained by experimental results on the proxy dataset.
   
   Experiments on Imagenette and Imagewoof \citep{imagenette} show that running hyperparameter search on the easiest 10\% of examples explains 81\% of the variance in experiment results on the target task, and using the easiest 50\% of examples can explain 95\% of the variance, significantly more than training on all the data for fewer epochs, a more widely used baseline.
   These ``easy" proxies are higher quality than training on the full dataset for a reduced number of epochs (but equivalent computational cost), a technique used by state-of-the art AutoML algorithms, and, unexpectedly, higher quality than proxies constructed from the hardest examples. Without access to a trained model, researchers can improve proxy quality by restricting the subset to fewer classes; proxies built on half the classes are higher quality than those with an equivalent number of examples spread across all classes.
\end{abstract}

\section{Introduction}
Despite rapidly improving software and hardware, training machine learning models remains a costly and time-consuming task. Training a model on a large dataset with many classes such as ImageNet takes roughly a day. Choosing, for example between 7 different hyperparameter values would take roughly a week of GPU time. For researchers and developers seeking to quickly iterate on algorithmic approaches or search for hyperparameters, it can be impractical to wait this long for a model to train.  Additionally, cheaper experiments might reduce the cost of running large AutoML experiments, allowing a wider group of researchers to contribute to a growing field.
  
In this paper, we explore the idea of making such experiments cheaper by creating ``proxy datasets" that exhibit two desirable qualities:  (1) they should be relatively cheap to train on (2) the relationship between hyperparameters and accuracy of training on the proxy task should closely resemble the dynamics of the full dataset. More specifically, a set of parameters (here referring to hyperparameters, model architectures, or other algorithmic approaches controlled by the researcher) that boost validation accuracy over another set of parameters on the proxy dataset cause a proportional boost in validation accuracy on the full dataset. Similarly, if a set of parameters performs poorly on a proxy dataset, it should perform poorly on the full dataset and the reduction in accuracy compared with a ``good" set of parameters should be proportional between the proxy and full datasets. In many hyperameter search tasks all that matters is finding the \textbf{best} configuration, so it is especially important that running hyperparameter search on a proxy dataset produces hyperparameters that are nearly as good as the true best hyperparameters.

In section 2, we discuss related work, which is more focused on the accuracy of training on a reduced dataset than the proxy quality of the reduced dataset. Section 3 discusses our dataset,  approach for generating proxy datasets, experimental settings, and metrics for measuring proxy quality. Section 4 presents results.

\section{Related Work}

\textbf{Some training examples have negative value.} In ``What is the value of a training example", \citep{value} propose to rank each training example by the validation accuracy of a model trained on that example and \textbf{all} examples from other classes. They run experiments on Pascal-VOC using SVMs, and find that they can create smaller subsets that outperform training on the full dataset. Using about 90\% of the original examples works best in their setting,  suggesting that some potentially mislabeled examples have negative value.

In ``An investigation of catastrophic forgetting in Neural Networks", \citep{forgetting} examine how often a neural network ``forgets" the correct prediction for an example, that is, how often it predicts the correct class for an example in one epoch and then makes an incorrect prediction in a subsequent epoch. In experiments on CIFAR-10\footnote{\citep{CIFAR10}}, they find that removing the 30\% of examples that have been forgotten the least after each epoch, does not impact validation accuracy, whereas removing a random 30\% of examples causes a much more significant reduction in accuracy. They also find that the same examples are forgettable and unforgettable for different ResNet architectures. We do not generate proxy datasets based on forget frequency, but instead use the loss of a fully trained model to generate ``easy" and ``hard" subsets. This choice is based on our hypothesis that frequently forgotten examples will tend to have higher loss at the end of training.

\textbf{Importance Sampling} Another group of works use approximations of the per-example gradient norm to estimate the training value of each example, with the insight that cause smaller weight updates are less valuable. In ``Not All Samples Are Created Equal: Deep Learning with Importance Sampling", to take one example, \citep{katharopoulos2018not} present a new trick for approximating example value and show that the benefit of skipping back propagation on low value examples can outweigh the cost of approximating the per-example gradient norm. Since an example with a low gradient norm likely has a low loss, we hypothesize that this technique is similar to our training technique, but do not explicitly test that assumption. There is also a large but somewhat older group of papers, on coreset construction, summarized by \citep{coreset}. These works focus on generating datasets with the same mean and standard deviation as the full dataset, sometimes by selecting examples but also, occasionally, by averaging them with K-Means.

\textbf{Dataset Distillation}, by \citep{distillation}, compresses a dataset to one example per class, with the goal of training that model to reasonable accuracy with 1 gradient descent step per example, by optimizing synthetic inputs such that a neural network trains well on them.
The authors show that AlexNet\footnote{\citep{AlexNet}} trained on distilled data achieves 54\% accuracy on CIFAR-10 if the distilllation process is given access to AlexNet's weight initialization, and 36\% without access.
On Imagenette, with AlexNet and Kaiming initialization\footnote{\citep{resnet}} training on distilled examples achieved 37\% accuracy, but we could not achieve results better than 20\% (10\% is random performance) with images larger than 32x32 or a ResNet architecture.

More importantly,  even with the best 37\% setup,  the distilled dataset turned out to be a poor proxy for the full dataset. Many changes that improved the default AlexNet implementation, like increasing the learning rate or using label smoothing loss, reduced proxy accuracy to random levels or caused gradient explosions. We interpret this result as evidence that (a) the images produced by dataset distillation are directly optimized to produce large gradients and should not be shown to the model multiple times and (b) distilled examples generated with a model that uses one set of hyperparameters do not transfer well to a model with different hyperparameters, as the original authors discuss.

To summarize our contribution, much previous work aims to reduce a dataset to a smaller dataset, either during or before training, and achieve good validation accuracy in less time. Our goal is not to train a strong final model on the smaller dataset, but rather to use the smaller dataset as a tool to accelerate hyperparameter search.\footnote{Our results could also be used with smarter hyperparameter search methods like Neural Architecture Search and Bayesian Optimization.\footnote{\citep{zoph_nas} and \citep{hyperpopt}}}

\section{Approach}
\subsection{Dataset}
Due to the large number of experiments required to demonstrate the statistical power of proxy results, we create proxy datasets on Imagenette and Imagewoof, two ImageNet proxies created by Jeremy Howard \footnote{\url{github.com/fastai/imagenette}}. Imagenette contains ImageNet examples from 10 easy to distinguish classes, while Imagewoof, a harder dataset for classification, contains data from 10 hard to distinguish classes -- different dog breeds. We run all experiments on 128x128 images, besides synthetic images generated by dataset distillation, which are 32x32.
We ran 36 different hyperparameter configurations x 2 datasets x 6 proxy creation strategies x 3 time budget levels, for a total of 1,296 training runs.

\subsection{Experiments}

\textbf{Hyperparameter Configurations (defaults in bold)}:  

\begin{enumerate}
    \item Architecture: \textbf{XResnet50}, XResnet18 or XResnet101.
    \item Learning Rate: $[.001,  \textbf{.003}, 0.007, .01, .1]$
    \item Stem Channels: How many channels for to output from each of ResNet's first two convolutional layers. This is two parameters set independently [[4, \textbf{32}, 48],  [4, \textbf{32}, 48]]
    \item Flip LR Probability: [.0, .25, \textbf{.5}] This parameter controls data augmentation. Specifically, how often to flip a training image horizontally.  
    \item Optimizer: {\textbf{Adam}, SGD, RMSProp}
    \item Loss Function: Cross Entropy Loss \textbf{with} or without label smoothing. Label Smoothing \citep{labelsmoothing} attempts to decrease a model's sensitivity to mislabeled examples by modifying the cross entropy loss function to use target values of 0.9 for the labeled class instead of instead of 1.0.\footnote{Label Smoothing seems to degrade performance slightly on our experiments, likely because they don't use mixup.}
\end{enumerate}

We do not run every possible combination of these hyperparameter values. Instead, we change one parameter at a time, while leaving the others as their default (bolded) values in order to increase the diversity of our search space. Table \ref{best_pars} shows the best performing hyperparameters on each dataset, which are slightly different than the defaults.

\begin{table}[!htbp]
\centering
\begin{tabular}{@{}lll@{}}
\toprule
                 & Imagenette   & Imagewoof  \\ \midrule
Architecture             & XResnet50 & XResnet50 \\
Label Smoothing & \textbf{False}       &\textbf{ False}       \\
Learning Rate               &\textbf{ 0.007 }      & 0.003       \\
Flip LR Probability      & 0.5         & \textbf{0.25 }       \\

Stem-channels-1  & 32.0        & 32.0        \\
Stem-channels-2  & 32.0        & 32.0        \\
Optimizer              & Adam        & Adam        \\ \midrule
Val Accuracy         & 0.926       & 0.801       \\
Val Accuracy for Default Params         & 0.920       & 0.784       \\
\bottomrule
\end{tabular}
\caption{Best hyperparameters for the target task-- differences from default are \textbf{bold}.}
\label{best_pars}
\end{table}

\textbf{Proxy Creation Strategies}
\begin{enumerate}
    \item All the Classes + Random Sampling: this is our baseline.
    \item Half the Classes + Random Sampling
    \item Easiest Examples: Both easy and hard examples are selected by training a model with the default hyperparameters on the full dataset then using that model to evaluate the loss on each example. Low loss examples are considered easy. 
    \item Hardest Examples: High loss examples selected using the same procedure.
    \item Fewer Epochs: train models for {1, 5, 10,} epochs. We refer to this as a proxy creation strategy because it reduces the cost of obtaining an experimental result, and can theforefore be thought of as a proxy for the target task.
\end{enumerate}

\textbf{Experimental Settings}
For each hyperparameter setting and proxy creation strategy (including the baseline -- use all the data), we train for the required epochs with the relevant hyperparameter setting. The validation data is always the same 500 examples, besides for the half-classes proxy creation strategies, where we remove classes that are not shown in the training set. For each run, we take the best validation accuracy at any epoch as the proxy result.\footnote{This tends to increase top 1 Accuracy for those proxies but does not impact proxy quality metrics.}  All experiments are run with the One-Cycle learning rate schedule, from \citep{ OneCycle},  and half-precision training, using a slightly modified version of the \text{train imagenette} script from the fastai library.  Running the target task (20 epochs) on either dataset  takes about 30 minutes and costs 52 cents on a P100 GPU. 

\textbf{Measuring Proxy Quality}
We propose three metrics for evaluation the quality of a proxy.
First the $r^2$ of the regression 
$$TargetAcc[i] = \beta * ProxyAcc[i]$$ where $i$ denotes the $i$th hyperparameter configuration.
This is also the covariance of the Target Accuracy and Proxy Accuracy.

In order to produce $r^2$ statistics in a wider range we normalize all accuracies within each dataset.\footnote{This way simply achieving less accuracy on Imagewoof, a harder dataset, does not give a proxy positive $r^2$ } This metric can be interpreted as the amount of variance in target experiment results that can be explained by proxy experiment results. Higher is better.

The second metric is the Spearman correlation between Proxy and Target results, ignoring poorly performing hyperparameters.
$$SpearmanR(ProxyAcc[i_{good}], TargetAcc[i_{good}]$$  where $i_{good}$ is the best performing hyperparameter configurations for a given proxy creation strategy.\footnote{For example. if we ran 4 experiments and achieved [.92, .93, .84, .86] on the proxy, and [.95, .99, .81] on the Target Acc. The statistic would be Spearman([.92, .93], [.95, .99])}
This statistic is motivated by the insight that in most hyperparameter search use cases, the user mostly cares about the relative rankings of the best configurations, which is what this statistic measures.\footnote{We also tried to measure regret -- how the best hyperparameters found on the proxy task before on the target task, but found this metric unstable, even averaged across randomly sampled hyperparameter configurations.}

The third metric


\textbf{Cost Adjustment}
Since the $r^2$ of a proxy is very correlated with it's computational cost, we also measure Cost-Adjusted $r^2$, by extracting the residual of a second regression:
$$
 TargetAcc_{i} = \beta_0 + \beta_1 cost_{i} +  \beta_2 cost_{i}^2   + \beta_3 cost_{i}^3 +  \text{Cost-Adjusted-}r^2_{i}
$$
We chose to use three polynomial terms by adding terms until they were given 0 coefficient by Lasso Regression.\footnote{\url{https://scikit-learn.org/stable/modules/generated/sklearn.linear_model.LassoCV.html}} Visually, $\text{Cost-Adjusted-}r^2 _ {[i]}$ is the distance between each point $i$  and the line in Figure \ref{headline}.

\section{Results}

\begin{figure}
\centering

\includegraphics[width=79mm]{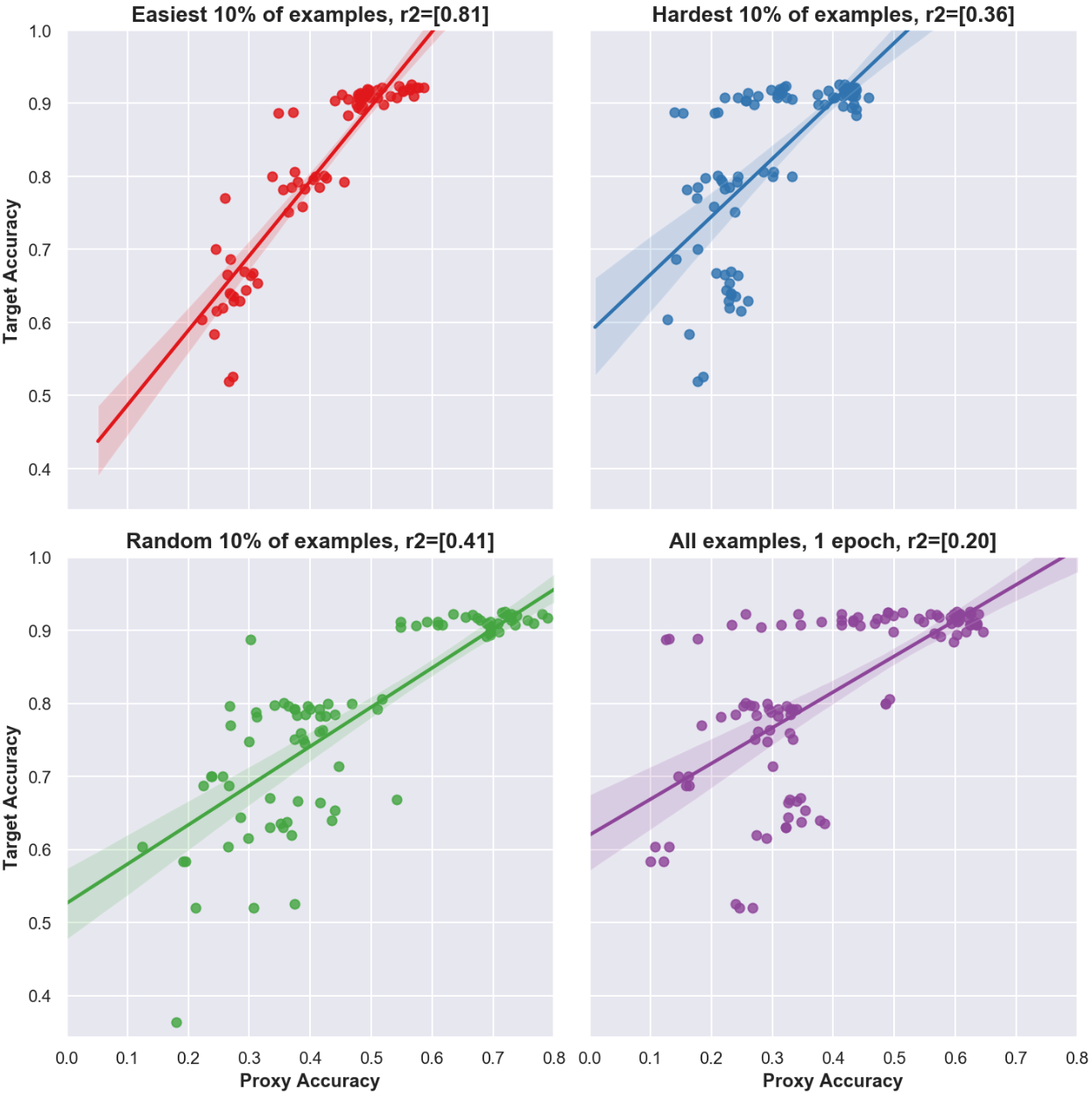}
\caption{These plots show the relationship between proxy accuracy and target accuracy for 4 different proxy creation strategies. Compared to the full task, the first three strategies take 10\% of the time to train, while the 1 epoch strategy takes 5\% of the time. The relationship for the easy example generation strategy (top right, red) is much stronger than for the blue (hard) example strategy, and green (random) example strategy, at equivalent computational cost.}
\label{scatters}
\end{figure}

\begin{figure*}
    \centering
    \figuretitle{Proxy Quality vs. Relative Cost}
    \begin{minipage}{.5\textwidth}
        \centering
        \includegraphics[width=60mm]{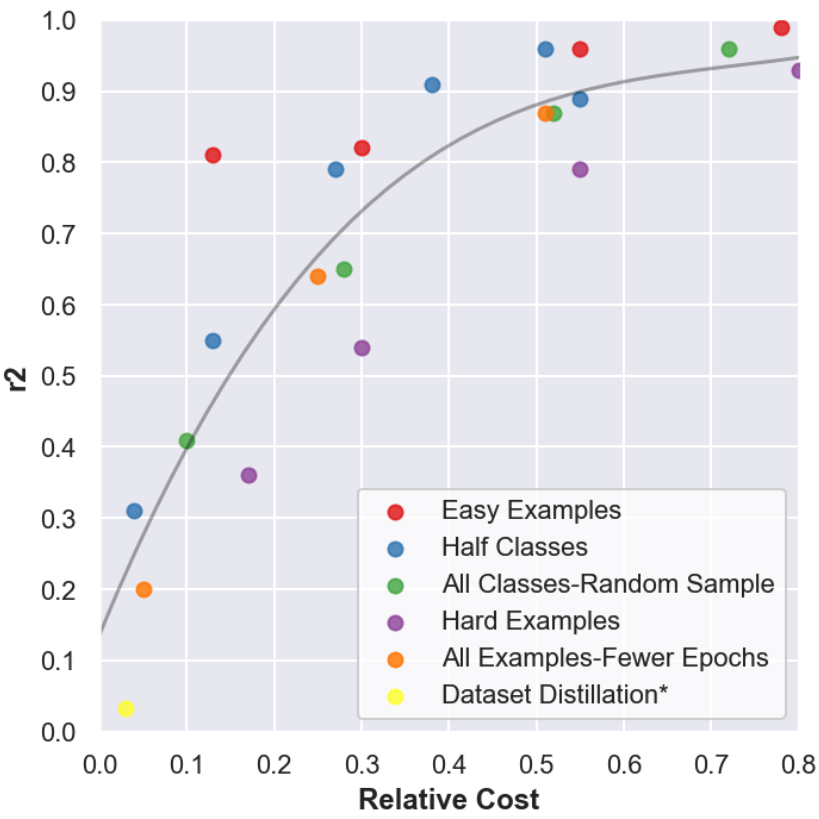}
   \end{minipage}%
   \begin{minipage}{.5\textwidth}
        \centering
        \includegraphics[width=60mm]{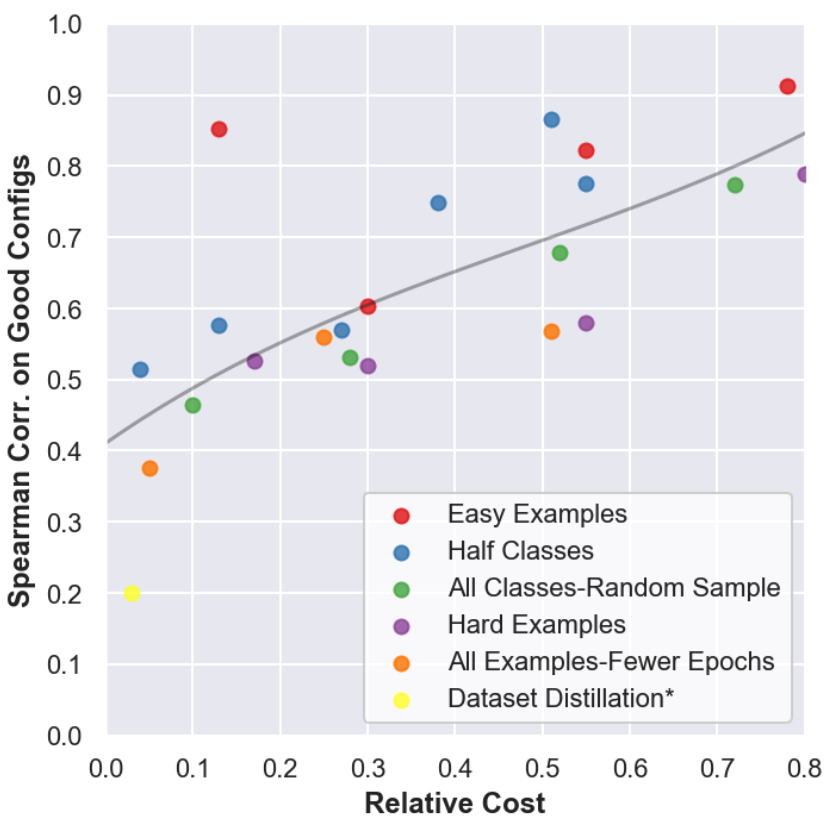}
   \end{minipage}
    \caption{Proxy quality vs. relative cost for different sampling strategies. Two measures of proxy quality are shown: $R^2$ (left) and Spearman correlation coefficient (right) for ``good configurations" (those for which the model trains above a minimum accuracy). The distance between a point and the line can be interpreted as cost-adjusted value of a specific proxy creation strategy. Relative Cost= $\frac{Proxy Time}{Target Time}$, where target time (how long it takes to train on the full dataset) is roughly 30 minutes.}
    \label{headline}
\end{figure*}

\textbf{General Trends}  Figure \ref{headline}
shows our two metrics of proxy quality for all proxy creation strategies against the computation cost of the proxy.  As expected, costlier proxies tend to be more correlated to the target than cheaper ones, but some cheaper proxies perform reasonably well. With the easiest 10\% of examples as a proxy, one can explain 81\% of the variance in target outcomes. 

On 50\% of the data two strategies -- \textit{Easy Examples} and \textit{Half Classes} have \textgreater 95\% $r^2$, and these two strategies beat the others across different Computational costs (the X axis).  All of these strategies recover the same best hyperparameters as the target task.

Figure \ref{scatters} plots proxy performance vs target performance for 4 different proxies that all take less than 3 minutes to train (10\% of the target task). Each data point in each chart represents the proxy and target results for some hyperparameter setting. The easy-example (red) strategy's regression line has an $r^2$ of 81\% of the variance in target results, much more than the 20\% $r^2$ from using the 10\% hardest examples.

\textbf{Hard vs Easy}
Proxying by selecting the easiest examples is the best performing strategy as shown in Figure 1. Another way to make the task easier, sampling half of the classes, is the second best proxy creation strategy, only slightly underperforming direct selection of easy examples on a cost-adjusted basis. These two strategies: restricting to the easiest examples, and restricting to half the classes before randomly sampling, outperform the other proxy creation strategies at all cost levels. Random Sampling and Training for fewer epochs perform almost equivalently to each other, but worse than the easier strategies. Selecting only hard examples is the worst strategy. This result is not caused by mislabeled or unlearnable examples; discarding the 5\% hardest examples, helps the 50\% hardest example proxy by 2\% while reducing the computational cost, but that proxy still underperforms Random Sampling. 
These trends continue in Figure 2, which plots our second metric, the Spearman correlation between proxy result and target result for good hyperparameter configurations.

At first glance, these results seems to contradict the Forgetting paper's evidence that easy/rarely forgotten examples can be discarded during training without event, but given the fairly small datasets we are working with, and the fact that our models never get to see the easy examples before they are discarded, 
we hypothesize that hard examples are much more useful later in the training process, after the model has already learned the basic features of the dataset from the easy examples. 

These ``easy" subsets are better proxies than training on the full dataset for a reduced number of epochs (but equivalent computational cost), and persist even if we remove the hardest 5\% of examples, which might be mislabeled.

Table \ref{expanded}, in the appendix, shows all metrics for all proxy creation strategies at different complexity levels.
\begin{figure}[!htbp]
\centering
\includegraphics[width=79mm]{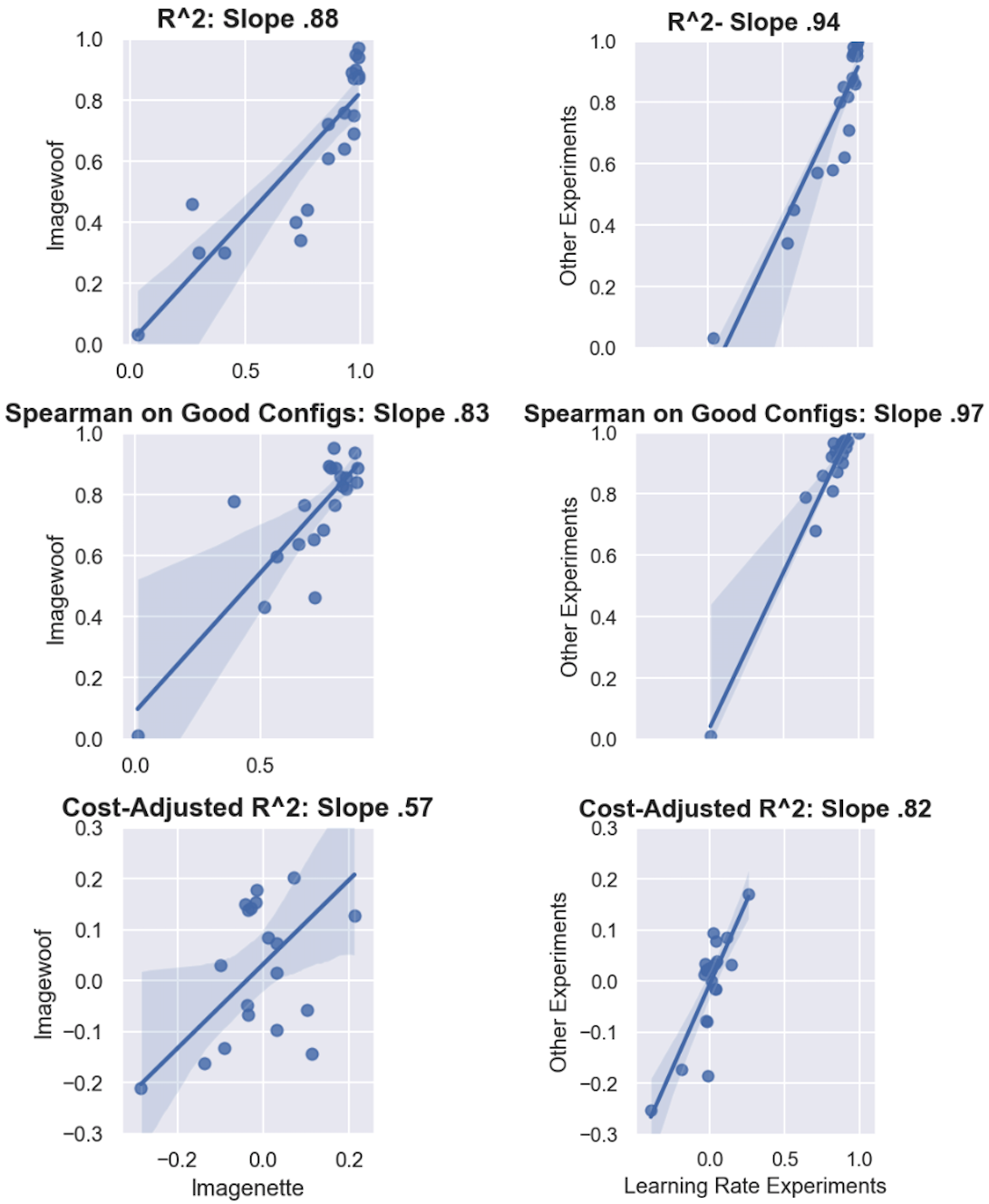}
\caption{Correlation between measurements across independent settings. Each row is one metric. The left column compares measurements for the same proxy strategy on Imagenette and Imagewoof. The right column compares different halves of the hyperparameter grid.}
\label{consistency}
\end{figure}

\textbf{Proxy Quality Transfers to New Settings} Figure \ref{consistency} shows that all three of our proxy quality measurements generalize between datasets and between independent hyperparameter grids.

On the left hand side we split our experimental results by dataset and then rerun metrics. By all three metrics, including the cost-adjusted $r^2$, proxy creation strategies that measure well on Imagenette tend to measure well on Imagewoof, and vice versa.  The right hand side splits the hyperparameter configurations in two roughly equally sized non-overlapping subsets -- configurations that change the learning rate (``Learning Rate Experiments") and those that change any other hyperparameter. Again, proxy creation strategies that measure well on learning rate experiments tend to measure well on other experiments, by all three metrics. Although this result is not plotted, the easiest example proxies continue to perform the best in all settings.

\begin{figure}
\figuretitle{Correlation between Intermediate and Final Accuracy}
\centering
\includegraphics[width=79mm]{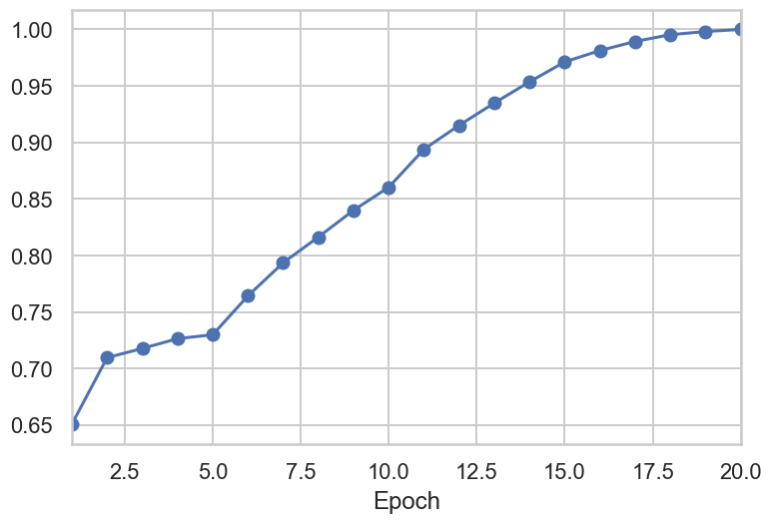}
\caption{The correlation between the accuracy after N epochs (X axis) and the final best validation accuracy for our 20 epoch models.}
\label{corrtime}
\end{figure}

Although we do not investigate this phenomenon as deeply, another way to reduce the cost of hyperparameter search might be to cancel underperforming models before they are done training and redirect the saved resources to other experiments. Figure \ref{corrtime} examines how much we know about the final model performance at a given epoch, and suggests that there appears to be a significant jump in information in the second epoch, and second, that late epochs continue to provide valuable information. These results are admittedly very dependent on our context -- if we trained for 30 epochs the curve would likely flatten out more quickly, but they still should inspire caution. If we randomly choose two runs that use all the data, but with different hyperparameters, the chance that the model with the higher validation accuracy after 1 epoch outperforms the other model is only 72\%. After 10 epochs this number is 81\%, and after 15 epochs it increases to 96\%. 

\section{Conclusion}
Our results suggest that hyperparameter search can be accelerated by using small subsets of the data. Running hyperparameter search on the easiest 10\% of examples explains 81\% of the variance in experiment results on the target task, and using the easiest 50\% of examples can explain 95\% of the variance, and all three of these strategies generate the same optimal hyperparameters as the target task. Proxy datasets built using the easiest examples are consistently higher quality than those built with hardest examples.  This pattern persists across datasets, and independent slices of the parameter grid.

\section{Future Work}
Imagenette and Imagewoof are both subsets of ImageNet, and it would be interesting to test whether the same high-level proxy quality patterns persist in other CV datasets, like CIFAR-10, or other domains like NLP. 
Additionally, it would be interesting to put these ideas into practice in the context of smarter hyperparameter selection systems and re-measure the speed/accuracy tradeoff. It would similarly be interesting to develop a regret style metric that actually measures the accuracy lost by taking shortcuts on a large enough set of experiments to establish consistency.
Finally, the failure of the hard example proxies suggests that they might be better used in a curriculum learning inspired approach, where we switch from easier to harder proxy sets, might lead to even better proxies.  Relatedly, initializing from pretrained weights might reduce the usefulness of easy examples.

\section{Appendix}
\subsection{Contributions}
As an experienced data scientist and machine learning engineer, Sam was the creative force behind much of the project design and direction, including deciding on sampling strategies, hyperparameter grids to run, and metrics to evaluate proxy quality. He also made major contributions to the experimental framework code, reproduced Dataset Distillation \citep{distillation}, bootstrapped the analysis code, and led the charge on the final report writeup.

Eric contributed by bootstrapping the experiment framework code and later enhancing it to support multiple sampling strategies, running experiments and collecting results, analyzing results, and creating plots.

\subsection{Acknowledgements}
We'd like to thank Jeremy Howard for inspiring us with his Imagenette \citep{imagenette} dataset to further explore the idea of creating proxy datasets, and Tongzhou Wang for helping us get Dataset Distillation running and quickly responding to our GitHub issues.

\begin{table*}[!htbp]
\title{Expanded Results}
\centering
\begin{tabular}{@{}llllll@{}}
\toprule
Strat and Size         & r2    & Spearman & Cost Adjusted R2 & Relative Cost & Proxy Creation Strategy \\ \midrule
All Classes-1.0        & 1.0   & 1.0      & 0          & 1.0           & Baseline                \\
hard-0.25-1.0          & 0.99  & 0.912    & 0.0556           & 0.7859        & Easy Examples           \\
Half Classes-1.0       & 0.96  & 0.866    & 0.0659           & 0.515         & Half Classes            \\
Other Half Classes-1.0 & 0.89  & 0.776    & -0.0142          & 0.5546        & Half Classes            \\
Half Classes-0.7       & 0.91  & 0.748    & 0.0739           & 0.3876        & Half Classes            \\
hard-0.05-0.5          & 0.79  & 0.605    & -0.1071          & 0.5257        & Hard Examples (*)       \\
hard-0.5-1.0           & 0.96  & 0.823    & 0.0553           & 0.5568        & Easy Examples           \\
hard-0.9-1.0           & 0.81  & 0.853    & 0.2922           & 0.1284        & Easy Examples           \\
All Classes-0.7        & 0.96  & 0.774    & 0.0325           & 0.728         & Random Sample           \\
hard-0.75-1.0          & 0.82  & 0.604    & 0.0556           & 0.2998        & Easy Examples           \\
Half Classes-0.1       & 0.31  & 0.515    & -0.0231          & 0.045         & Half Classes            \\
All Classes-0.5        & 0.87  & 0.679    & -0.0274          & 0.5268        & Random Sample           \\
hard-0.0-0.75          & 0.93  & 0.789    & -0.0088          & 0.8137        & Hard Examples           \\
hard-0.0-0.5           & 0.79  & 0.579    & -0.1147          & 0.5568        & Hard Examples           \\
Half Classes-0.5       & 0.79  & 0.57     & 0.0577           & 0.2698        & Half Classes            \\
All Classes-1.0-ep10   & 0.87  & 0.568    & -0.0238          & 0.5139        & Fewer Epochs            \\
All Classes-1.0-ep1    & 0.2   & 0.375    & -0.1489          & 0.0514        & Fewer Epochs            \\
Half Classes-0.25      & 0.55  & 0.576    & 0.0279           & 0.1307        & Half Classes            \\
All Classes-0.1        & 0.41  & 0.464    & -0.0516          & 0.1007        & Random Sample           \\
All Classes-0.25       & 0.65  & 0.532    & -0.0931          & 0.2794        & Random Sample           \\
All Classes-1.0-ep5    & 0.64  & 0.56     & -0.0773          & 0.257         & Fewer Epochs            \\
hard-0.0-0.25          & 0.54  & 0.519    & -0.2244          & 0.2998        & Hard Examples           \\
hard-0.0-0.1           & 0.36  & 0.526    & -0.2349          & 0.1713        & Hard Examples           \\
distillation           & 0.032 & 0.01     & -0.2515          & 0.0257        & distillation            \\ \bottomrule
\end{tabular}
\caption{The three metrics we use for each proxy creation strategy we evaluated.  Hard Examples (*) removes the hardest 5\% of examples. Each row of the dataset represents statistics computed over 72 different hyperparameter configurations.}
\label{expanded}
\end{table*}
\newpage
{\small
\bibliography{egbib}

\begin{thebibliography}{13}
\providecommand{\natexlab}[1]{#1}
\providecommand{\url}[1]{\texttt{#1}}
\expandafter\ifx\csname urlstyle\endcsname\relax
  \providecommand{\doi}[1]{doi: #1}\else
  \providecommand{\doi}{doi: \begingroup \urlstyle{rm}\Url}\fi

\bibitem[Bachem et~al.(2017)Bachem, Lucic, and Krause]{coreset}
Olivier Bachem, Mario Lucic, and Andreas Krause.
\newblock Practical coreset constructions for machine learning.
\newblock \emph{arXiv preprint arXiv:1703.06476}, 2017.

\bibitem[Bergstra et~al.(2013)Bergstra, Yamins, and Cox]{hyperpopt}
James Bergstra, Dan Yamins, and David~D Cox.
\newblock Hyperopt: A python library for optimizing the hyperparameters of
  machine learning algorithms.
\newblock In \emph{Proceedings of the 12th Python in science conference}, pages
  13--20. Citeseer, 2013.

\bibitem[Goodfellow et~al.(2013)Goodfellow, Mirza, Xiao, Courville, and
  Bengio]{forgetting}
Ian~J Goodfellow, Mehdi Mirza, Da~Xiao, Aaron Courville, and Yoshua Bengio.
\newblock An empirical investigation of catastrophic forgetting in
  gradient-based neural networks.
\newblock \emph{arXiv preprint arXiv:1312.6211}, 2013.

\bibitem[He et~al.(2016)He, Zhang, Ren, and Sun]{resnet}
Kaiming He, Xiangyu Zhang, Shaoqing Ren, and Jian Sun.
\newblock Deep residual learning for image recognition.
\newblock In \emph{Proceedings of the IEEE conference on computer vision and
  pattern recognition}, pages 770--778, 2016.

\bibitem[Howard(2019)]{imagenette}
Jeremy Howard.
\newblock Imagenette.
\newblock Github repository with links to dataset, 2019.
\newblock https://github.com/fastai/imagenette.

\bibitem[Katharopoulos and Fleuret(2018)]{katharopoulos2018not}
Angelos Katharopoulos and Fran{\c{c}}ois Fleuret.
\newblock Not all samples are created equal: Deep learning with importance
  sampling.
\newblock \emph{arXiv preprint arXiv:1803.00942}, 2018.

\bibitem[Krizhevsky and Hinton(2009)]{CIFAR10}
Alex Krizhevsky and Geoffrey Hinton.
\newblock Learning multiple layers of features from tiny images.
\newblock Technical report, Citeseer, 2009.

\bibitem[Krizhevsky et~al.(2012)Krizhevsky, Sutskever, and Hinton]{AlexNet}
Alex Krizhevsky, Ilya Sutskever, and Geoffrey~E Hinton.
\newblock Imagenet classification with deep convolutional neural networks.
\newblock In \emph{Advances in neural information processing systems}, pages
  1097--1105, 2012.

\bibitem[Lapedriza et~al.(2013)Lapedriza, Pirsiavash, Bylinskii, and
  Torralba]{value}
Agata Lapedriza, Hamed Pirsiavash, Zoya Bylinskii, and Antonio Torralba.
\newblock Are all training examples equally valuable?
\newblock \emph{arXiv preprint arXiv:1311.6510}, 2013.

\bibitem[Pereyra et~al.(2017)Pereyra, Tucker, Chorowski, Kaiser, and
  Hinton]{labelsmoothing}
Gabriel Pereyra, George Tucker, Jan Chorowski, {\L}ukasz Kaiser, and Geoffrey
  Hinton.
\newblock Regularizing neural networks by penalizing confident output
  distributions.
\newblock \emph{arXiv preprint arXiv:1701.06548}, 2017.

\bibitem[Smith and Topin(2018)]{OneCycle}
Leslie~N Smith and Nicholay Topin.
\newblock Super-convergence: Very fast training of residual networks using
  large learning rates.
\newblock 2018.

\bibitem[Wang et~al.(2018)Wang, Zhu, Torralba, and Efros]{distillation}
Tongzhou Wang, Jun-Yan Zhu, Antonio Torralba, and Alexei~A Efros.
\newblock Dataset distillation.
\newblock \emph{arXiv preprint arXiv:1811.10959}, 2018.

\bibitem[Zoph and Le(2016)]{zoph_nas}
Barret Zoph and Quoc~V Le.
\newblock Neural architecture search with reinforcement learning.
\newblock \emph{arXiv preprint arXiv:1611.01578}, 2016.

\end{thebibliography}
}
\end{document}